\def\year{2021}\relax
\documentclass[letterpaper]{article} 
\usepackage{aaai21}  
\usepackage{times}  
\usepackage{helvet} 
\usepackage{courier}  
\usepackage[hyphens]{url}  
\usepackage{graphicx} 
\usepackage{natbib}  
\usepackage{caption} 
\usepackage[switch]{lineno}
\usepackage{amsmath}
\usepackage{physics}
\usepackage{breqn}
\usepackage{subfigure}
\usepackage{multirow}
\usepackage{tabularx}
\title{A Spatial-Temporal Graph Based Hybrid Infectious Disease Model with Application to COVID-19}

\author {

        Yunling Zheng,\textsuperscript{\rm 1}
        Zhijian Li, \textsuperscript{\rm 1}
        Jack Xin \textsuperscript{\rm 1}
        Guofa Zhou \textsuperscript{\rm 2}\\
}
\affiliations {
    \textsuperscript{\rm 1} Department of Mathematics, UC Irvine \\
    \textsuperscript{\rm 2} College of Health Sciences, UC Irvine \\
}

\begin{document}
\maketitle

\begin{abstract}

As the COVID-19 pandemic evolves, reliable prediction plays an important role for policy making. The classical infectious disease model SEIR (susceptible-exposed-infectious-recovered) is a compact yet simplistic temporal model. The data-driven machine learning models such as RNN (recurrent neural networks) can suffer in case of limited time series data such as COVID-19. In this paper, we combine SEIR and RNN on a graph structure to develop a hybrid spatio-temporal model to achieve both accuracy and efficiency in training and forecasting. We introduce two features on the graph structure: node feature (local temporal infection trend) and edge feature (geographic neighbor effect). For node feature, we derive a discrete recursion (called I-equation) from SEIR so that gradient descend method applies readily to its optimization. For edge feature, we design an RNN model to capture the neighboring effect and regularize the landscape of loss function so that local minima are effective and robust for prediction.  The resulting hybrid model (called IeRNN) improves the prediction accuracy on state-level  COVID-19 new case data from the US, out-performing standard temporal models (RNN, SEIR, and ARIMA) in 1-day and 7-day ahead forecasting. Our model accommodates various degrees of reopening and provides potential outcomes for policymakers.

\end{abstract}

\section{Introduction}
The classical infectious disease model, the SEIR model \cite{hethcote_2000}, is a variation on the basic SIR model \cite{anderson_may_1992}. It assumes that all individuals in the population can be categorized into one of the four compartments:  Susceptible, Exposed, Infected and Removed, during the period of pandemic. The model describes the evolution of the compartmental populations in time by a system of nonlinear ordinary differential equations (ODE):
\begin{eqnarray*}
  \dv{S}{t} & = & - \beta_1 S I \\
  \dv{E}{t} & = & \beta_1 S I  - \sigma_1 E \\
  \dv{I}{t} & = & \sigma_1 E - \gamma I\\
  \dv{R}{t} & = & \gamma I\\
\end{eqnarray*}
The total population $S+E+I+R$ is invariant in time, which we shall normalize to 1 or 100 \%  in the rest of this paper. 
Clearly, SEIR is a simplistic temporal model of a given region or country. 

However, the infectious disease data often provides not only temporal but also spatial information as in the case of COVID-19, see \cite{us_data}. A natural idea is to elevate SEIR model to a spatio-temporal model so that it can be trained from the currently reported data and make more accurate real-time prediction. 
See \cite{roosa_2020} for temporal modeling on cumulative cases of China and real-time prediction.

In this paper, we set out to model the latent effect of inflow cases from the geographical neighbors to capture spacial spreading effect of infectious disease. For the practical reason that the inflow data is not observable, machine learning methods such as regression and neural network are more suitable. 
As widely adopted in time-series prediction problem, linear statistical models, such auto-regressive model (AR) and its variants are standard methods to forecast time-series data with some distribution assumptions on the time series. And the Long Short Term Memory neural networks model (LSTM)  \cite{hochreiter1997long} for the natural language processing problem, can be applied to time series data, especially disease data. 
With additional spatial information, the graph-structured LSTM models show a better performance on spatio-temporal data. See the application to influenza data \cite{li_2019}\cite{deng2019graph} and crime and traffic data \cite{wang,wang2018graph}. 
However, such neural network models have a demand for a large training data to optimize the high dimensional parameters.
Yet the reliable daily data of COVID-19 in the US begins after March 2020 and limits the temporal resolution.  Applying space-time LSTM models \cite{li_2019,wang2018graph} directly to COVID-19 may lead to overfitting. 
In light of the shortage of data of COVID-19, we shall derive a hybrid SEIR-LSTM model with much fewer parameters than space-time LSTMs \cite{LaiCYL17}\cite{Wu2018DeepLF}.

\section{Related work}
In \cite{yang2015accurate}, ARGO (AutoRegression with Google search trends), a variant of AR, uses the google search trends to generate external feature of ARGO and forecasts influenza data from Centers for Disease Control of U.S.(CDC). ARGO is a linear statistical model that combines historical observations and external features. The prediction of influenza activity level is given by:

\[\hat{y}_t=u_t+\sum_{j=1}^{52}\alpha_jy_{t-j}+\sum_{i=1}^{100}\beta_iX_{i,t}.\]

where $\hat{y}_t$ is the predicted value at time $t$, and the optimization part of ARGO is:

\[\min_{\mu_y,\vec{\alpha},\vec{\beta}}\Big(y_t-u_t-\sum_{j=1}^{52}\alpha_jy_{t-j}-\sum_{i=1}^{100}\beta_iX_{i,t}\Big)^2\]
\[+\lambda_a||\vec{\alpha}||_1+\eta_a||\vec{\beta}||_1+\lambda_b||\vec{\alpha}||_2^2+\eta_b||\vec{\beta}||_2^2\]

where $\vec{\alpha}=(\alpha_1,\cdots,\alpha_{52})$ and $\vec{\beta}=(\beta_1,\cdots,\beta_{100})$. 
$y_{t-j}$ ($ 1\leq j\leq 52$) are historical values of past 52 weeks and $X_{i,t}$ ($1\leq i\leq 100$) are the google search trend features at time $t$. The feature are generated by top 100 of most related trends to influenza from google search at each time.
The additional regularization terms to linear regression model helps ARGO optimize. The numerical experiment from \cite{yang2015accurate} shows a better performance than machine learning models such as LSTM, AR, and ARIMA.

The \cite{li_2019} introduces a graph structured recurrent neural network (GSRNN) to further improve the forecasting accuracy of CDC influenza activity level data. From CDC data, the USA is divided into 10 Health and Human
Services (HHS) regions to report influenza activity level. These 10 regions are described as a graph in GSRNN with nodes $v_1,\cdots,v_{10}$ and a collection of edges based on geographic neighbor relationship (i.e. $E=\{(v_i,v_j)|v_i, v_j \ \text{are adjacent}\}$, $E$ is the set of all edges). By comparing the average record of activity levels, the 10 HHS region nodes are divided into two groups by relatively active level, the high active group $\mathcal{H}$, and low inactive group $\mathcal{L}$. The two group leads to 3 types of edges between them, $\mathcal{L}-\mathcal{L}$, $\mathcal{H}-\mathcal{L}$, and $\mathcal{H}-\mathcal{H}$, where each edge type has a customized RNN, called edge-RNN, to generate the edge features. There are also two kinds of RNNs for each node group to combine the edge feature with historical values and output the final prediction. Suppose a fixed node $v\in \mathcal{H}$. The edge feature of $v$ at time $t$ are $e_{v,\mathcal{H}}^t$ and $e_{v,\mathcal{L}}^t$, which are generated by the average of historical values of neighbor nodes of $v$ in corresponding groups. The edge features are the input of the corresponding edge-RNN of each edge:

$$f_v^t=\text{edgeRNN}_{\mathcal{H}-\mathcal{L}}(e_{v,\mathcal{L}}^t), \ \  h_v^t=\text{edgeRNN}_{\mathcal{H}-\mathcal{H}}(e_{v,\mathcal{L}}^t)$$
Then, the outputs of edge-RNNs are fed into the node-RNN of group $\mathcal{H}$ together with the node feature of $v$ at time $t$, denoted as $v^t$, to output the prediction of the activity level of node $v$ at time $t+1$, or $y_{v}^{t+1}$:

$$y_{v}^{t+1}=\text{nodeRNN}_{\mathcal{H}}(v^t,f_v^t,h_v^t).$$

\section{Our approach: IeRNN model}
We propose a novel hybrid spatio-temporal model, named IeRNN, by combining LSTM \cite{hochreiter1997long} and I-equation on a graph structure. The I-equation is a discrete in time model derived from SEIR differential equations. It resembles a nonlinear regression model of time series.  The LSTM framework is applied to model the latent  geographical inflow of infections.  Our IeRNN model, comparing to  \cite{li_2019,wang,wang2018graph}, is much more compact.

\subsection{Derivation of I-equation from SEIR ODEs}
As a variation to SEIR model, we shall construct additional features $I_e$ and $E_e$ that reveal the inflow population of infectious and exposed individuals from neighboring regions. Then we augment the SEIR differential equations with $I_e$ and $E_e$ as:
\begin{eqnarray}
  \dv{S}{t} & = & - \beta_1 \, S \, I - \beta_2\, S\, I_e  \label{sode}\\
  \dv{E}{t} & = & \beta_1\, S \, I + \beta_2\,  S\, I_e - \sigma_1 \, E - \sigma_2\, E_e \;\;\; \label{eode}\\
  \dv{I}{t} & = & \sigma_1 \, E + \sigma_2 \, E_e - \gamma I  \label{iode}\\
  \dv{R}{t} & = & \gamma \, I  \label{rode}
  \end{eqnarray}
  \medskip
  
  It still follows that 
  \begin{equation}
  S + E + I + R =  1  \label{constant}
\end{equation}
by normalizing compartmentalized populations to percentages of total population.   
From (\ref{sode}) and (\ref{rode}), we have
\begin{eqnarray}
  R (t) & = & R (t_0) + \gamma \int_{t_0}^t I (\tau) d \tau \label{Req}\\
  S & = & S_0 \exp \left( - \int_{t_0}^t (\beta_1 I + \beta_2 I_e) d \tau \right) \label{Seq}
\end{eqnarray}

Substituting  (\ref{Req}),  (\ref{Seq}) and  (\ref{constant}) in 
(\ref{iode}), we have a closed I-equation:

\begin{dmath}
      \gamma I + \dv{I}{t} - \sigma_2 E_e =  \sigma_1 \left( 1 - I
  (t) - R (t_0) - \gamma \int_{t_0}^t I (\tau) d \tau  - S_0 \exp \left( -
  \int_{t_0}^t (\beta_1 I + \beta_2 I_e) d \tau \right) \right)
\end{dmath}
The above derivation holds for time dependent coefficients $\beta_i=\beta_i(t)$,   $i=1,2$. Let $E_e=\tau I_e$, and write  $\sigma_2 \tau$ as $\sigma_2$. By the  explicit Euler and $(P+1)$-term Riemann sum approximation, we have a discrete time recursion:

\begin{dmath}
  \gamma I_t + I_{t + 1} - I_t - \sigma_2 I_{e, t} = \sigma_1 \alpha -  \sigma_1 I_t - \gamma \frac{t - t_0}{P + 1}  \sum_{j = 0}^P I_{t - j} - S_0  \exp \left( - \frac{t - t_0}{P + 1}  \sum_{j = 0}^P \, (\beta_1 I)_{t - j} +  (\beta_2 I_{e})_{t - j} \right)
\end{dmath}
which gives the {\it I-model}:

\begin{dmath}
  I_{t + 1} =  \sigma_1 \alpha + (1 - \sigma_1 - \gamma) I_t + \sigma_2 I_{e, t} - \gamma \frac{t - t_0}{P + 1}  \sum_{j = 0}^P I_{t - j} - S_0 \exp \left( - \frac{t - t_0}{P + 1}  \sum_{j = 0}^P\,  (\beta_1 I)_{t - j} + (\beta_2 I_{e})_{t - j} \right)
\label{eq:Ie}
\end{dmath}
If $I_{e}\equiv 0$ in I-model  (\ref{eq:Ie}), we get an approximation of the $I_t$ component of SEIR model, 
%
a nonlinear regression 
model in time for a single region, named the {\it I-equation}.

\begin{figure}[htbp]
    \centering
    \includegraphics[width=0.45\textwidth]{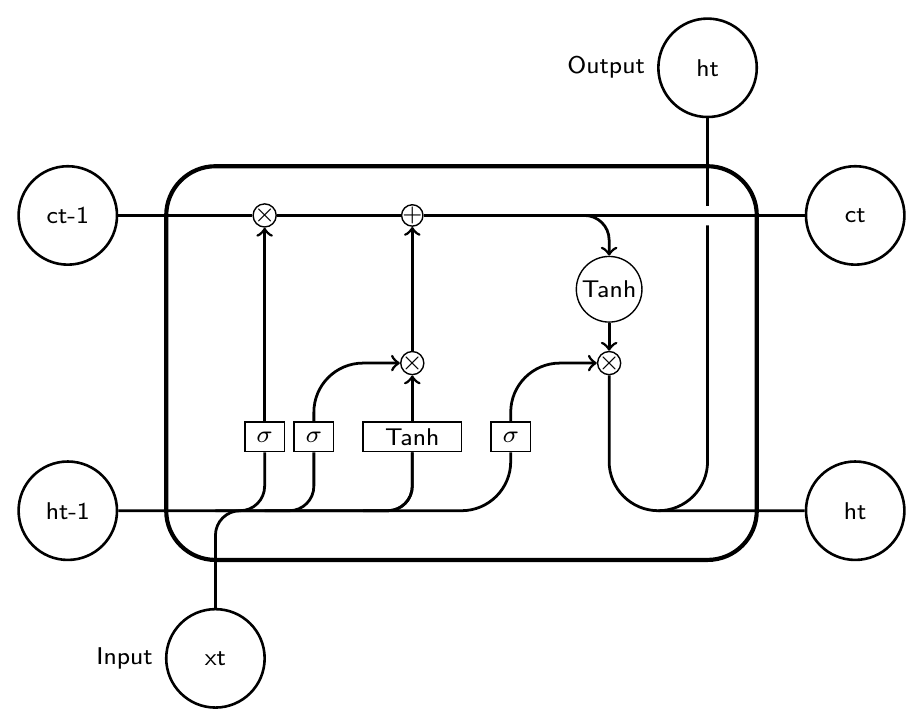}
    \caption{LSTM cell.}
    \label{fig:lstm}
\end{figure}

Since it is difficult to keep track of infectious and exposed populations  coming from neighboring regions, here we model $I_{e,t}$ as a latent feature in absence of a  mathematical formula or equation. To represent the latent feature from time-varying influx of infectious individuals, we make use of LSTM, a recurrent form of neural networks, see Fig. \ref{fig:lstm}.
\subsection{Generate edge feature with $I_e$}
\begin{figure}[htbp]
    \centering
    \includegraphics[width=0.5\textwidth]{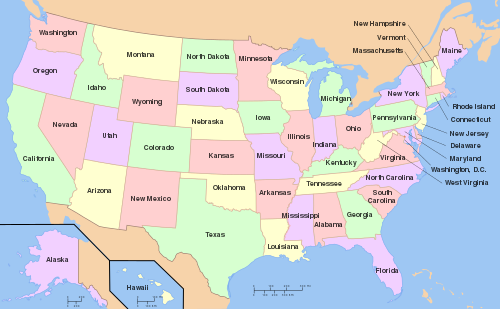}
    \caption{USA state map.}
    \label{fig:us_map}
\end{figure}

The spatial information based on US states map \cite{us_map}, see Fig. \ref{fig:us_map}, is formulated as an adjacent matrix $G=(g_{i,j})$. If two states $v_i$, $v_j$ are neighbors to each other, then $g_{i,j}=1$ otherwise is zero. With the variables of graph information, we can define the edge feature of state $v_i$ at time $t$:

$$
f_{i,t} = \frac{1}{\sum_{j} g_{i,j}}\,  \sum_{j} \, \left (g_{i,j}\,  \sum_{k=1}^{p}\,  I_{j,t-k}\right )
$$
where $I_{j,t}$ is the infectious population percentage in state $v_j$ at time $t$. 

\begin{figure}[htbp]
    \centering
    \includegraphics[width=0.45\textwidth]{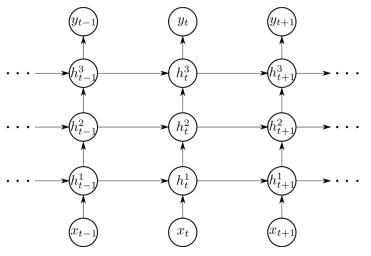}
    \caption{Stacked LSTM cells in edge-RNN.}
    \label{fig:stack_lstm}
\end{figure}

\begin{figure}[htbp]
    \centering
    \includegraphics[width=0.45\textwidth]{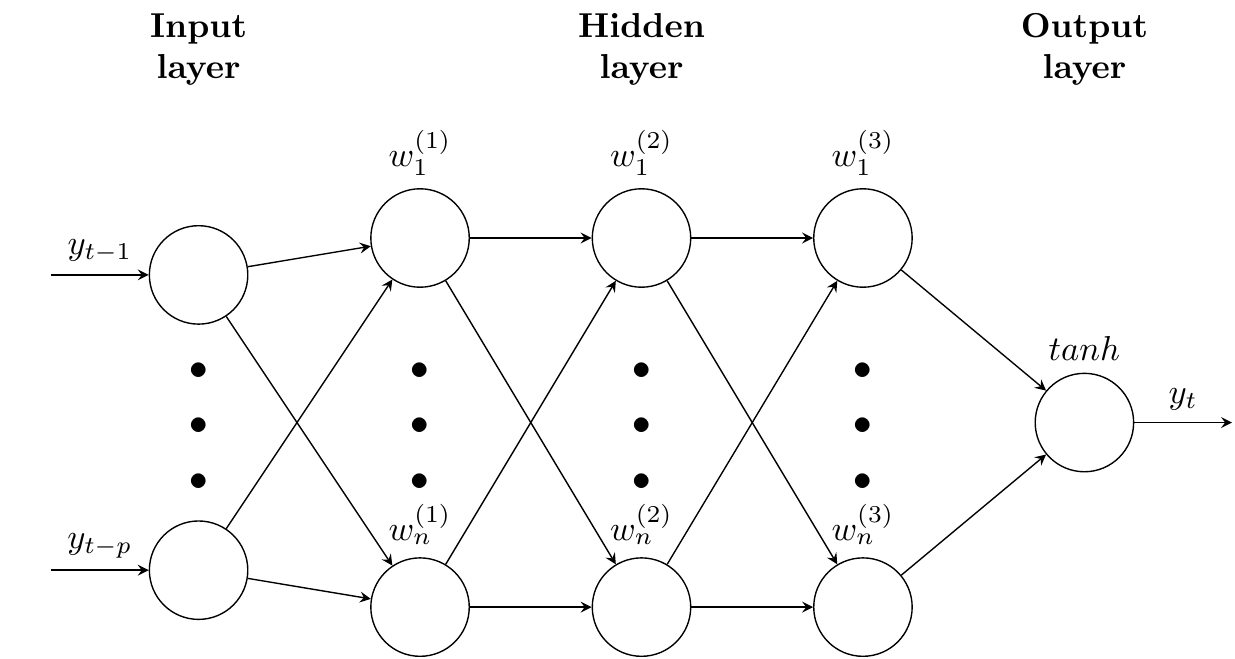}
    \caption{Fully connected dense layer.}
    \label{fig:fcnn}
\end{figure}

\begin{figure}[htbp]
    \centering
    \includegraphics[width=0.5\textwidth]{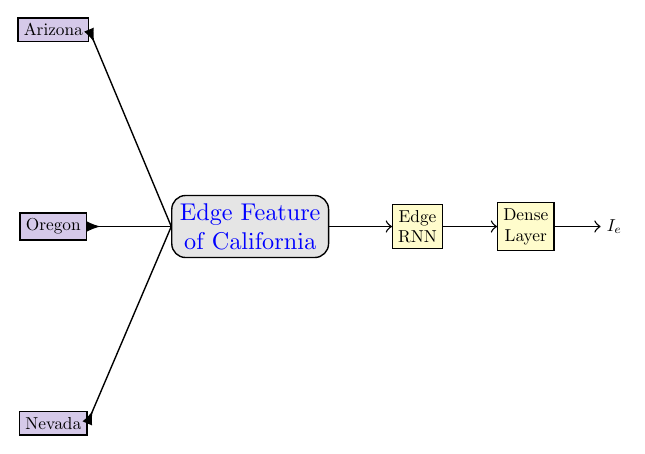}
    \caption{Generate $I_e$ of California.}
    \label{fig:edge_rnn}
\end{figure}

Then we design an edge-RNN composed of stacked LSTM cells Fig. \ref{fig:stack_lstm} with a following dense layer Fig. \ref{fig:fcnn} to output $I_e$. The edge feature $f_{i,t}$ is the input of the edge-RNN. The integrated procedure to generate edge feature is illustrated by Fig. \ref{fig:edge_rnn}, taking California as example. Our model, IeRNN, is named by this design of edge-RNN for $I_e$ and the I-equation:

\begin{dmath}
  I_{e,t} = \textit{Dense-Layer}(\textit{edge-RNN}(f_{i,t}))
\end{dmath}

\subsection{Policy response modeling}
During an epidemic, the rate of infection could change as governments start responding to the epidemic. The infectious rate would start decreasing due to the restrictive policy (partial or full lock-down) being put in place. 
We model the policy response by changing the parameter, $\beta_1$, in the ODE set (\ref{eq:Ie}).
By multiplying a control factor $\beta_{decay}$ (called test decay) to $\beta_1$, we can control  the infection levels in the future resulting from various degrees of opening policies.

The policy response $\beta_1$ is a function of time 
\cite{Li2020} 
to reflect no measure, restricting mass gatherings, reopening, lock-down for different times:

\begin{dmath}
  \beta_1(t) = \frac{2}{\pi} \arctan \left (\frac{-b (t-a)}{20}\right ) + 1 + c \, \exp\left (-\frac{(t-t_0)^2}{s}\right )
  \label{eq:beta}
\end{dmath}
where parameters $a$, $b$, $c$, $t_0$, $s$ are learned from fitting historical data.

\section{Experiment}

We use the COVID-19 data in United States\cite{us_data} to evaluate our IeRNN model in training and testing. 
From \cite{us_data}, we find the state level infectious data in the US.
Due to the incomplete recovered cases of US, we use the difference of cumulative cases in each state as the daily infected population. Then we use the population of US \cite{us_pop} to calculate the infectious rate of each state, where we assume the population of a state is constant during the period we concerned with. The data is split into training set  (133 days) and testing set (35 days) for model evaluation.

The loss function for training is mean squared error (MSE) of the output of model and true data value:

$$
  {\rm loss} = \frac{1}{T+1} \sum_{t=0}^{T}(I_t-\hat{I}_t)^2
$$
where the output of model $\hat{y}_t$ has the form (adapted from (\ref{eq:Ie})):

\begin{dmath}
  \hat{I}_t =  \sigma_1 \alpha + (1 - \sigma_1 - \gamma) I_{t-1} + \sigma_2 I_{e, t-1} - \gamma \frac{t - t_0}{P + 1}  \sum_{j = 0}^P I_{t-1-j} - S_0 \exp \left( - \frac{t - t_0}{P + 1}  \sum_{j = 0}^P\,  (\beta_1 I)_{t-1-j} + (\beta_2 I_{e})_{t-1-j} \right)
\label{eq:loss}
\end{dmath}
where we have parameters 
$(\alpha,\beta_1,\beta_2,\gamma,\sigma_1,\sigma_2)$.
Due to the interpretations of SEIR model, these parameter values  should range in the interval $[0,1]$.

We use gradient descent optimizer, Adam \cite{adam}, to train our IeRNN model. In each step, we update the weight of neural networks model (\ref{eq:Ie}) and the parameters of loss function (\ref{eq:loss}) separately with different length of step and regularization norms.

To assess the performance of our model, we design a series of numerical experiments to compare the IeRNN with I-equation, temporal LSTM and ARIMA. 

Regarding model size, the IeRNN and LSTM have about 4240 parameters while the I-equation and ARIMA have 5 parameters.


\subsection{Robustness in  parameter initialization}
\begin{figure}
\subfigure{
    \includegraphics[width=0.5\textwidth]{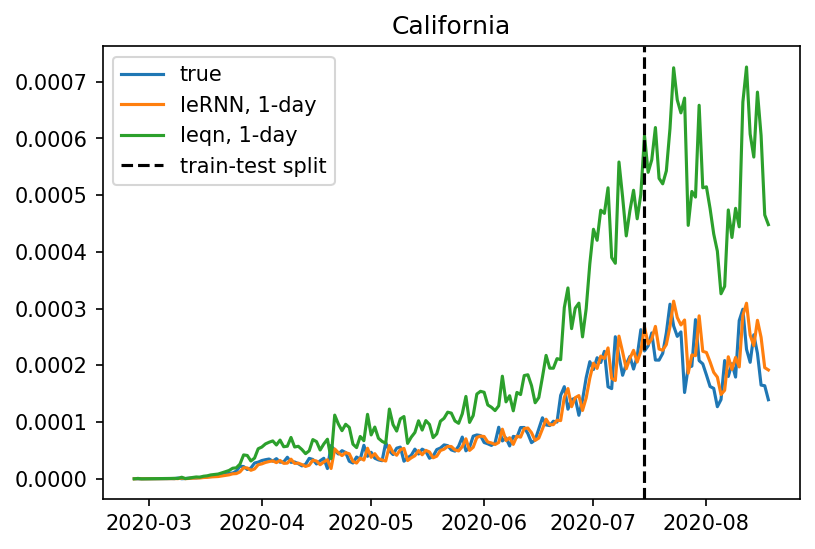}
}
\subfigure{
    \includegraphics[width=0.5\textwidth]{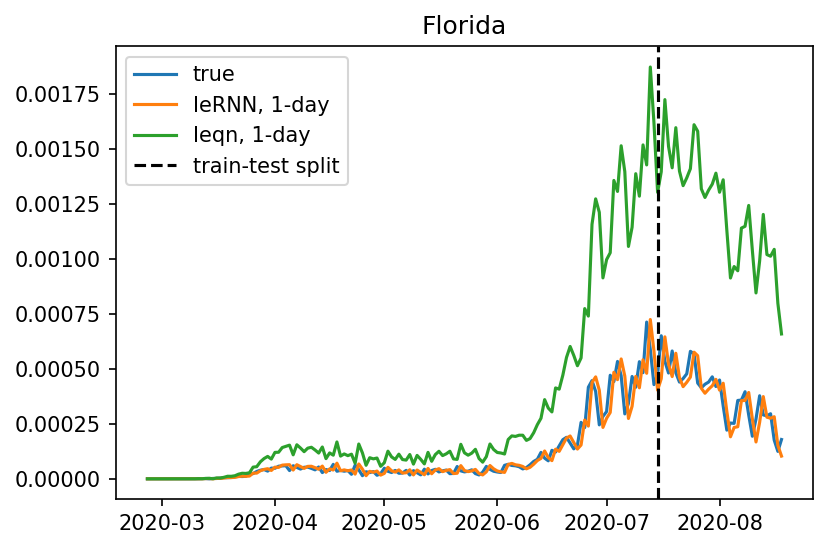}
}
\subfigure{
    \includegraphics[width=0.5\textwidth]{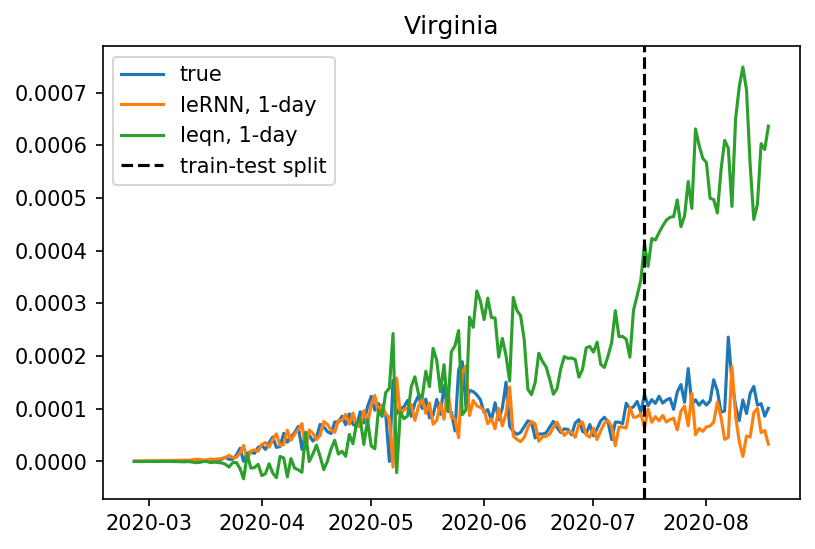}
}
\caption{Comparing 1-day ahead predictions of IeRNN and I-equation with training (testing) period to the left (right) of the vertical dashed line.  The vertical axis is fraction of newly infected people in the population. The horizontal axis is time in unit of days.}
\label{fig:1day}
\end{figure}

\begin{figure}
\subfigure{
    \includegraphics[width=0.5\textwidth]{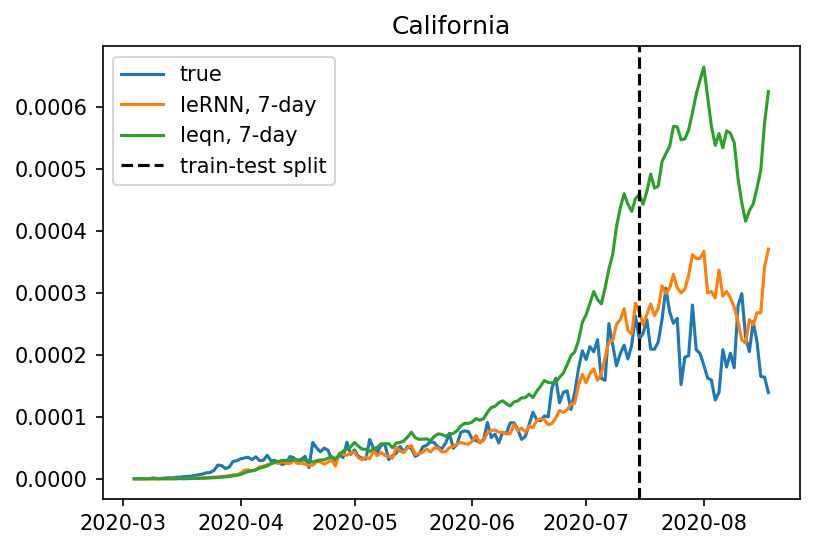}
}
\subfigure{
    \includegraphics[width=0.5\textwidth]{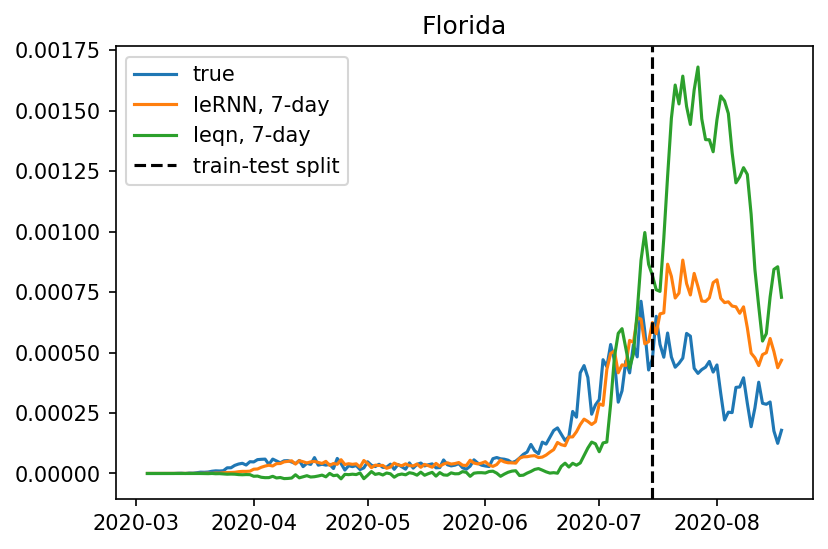}
}
\subfigure{
    \includegraphics[width=0.5\textwidth]{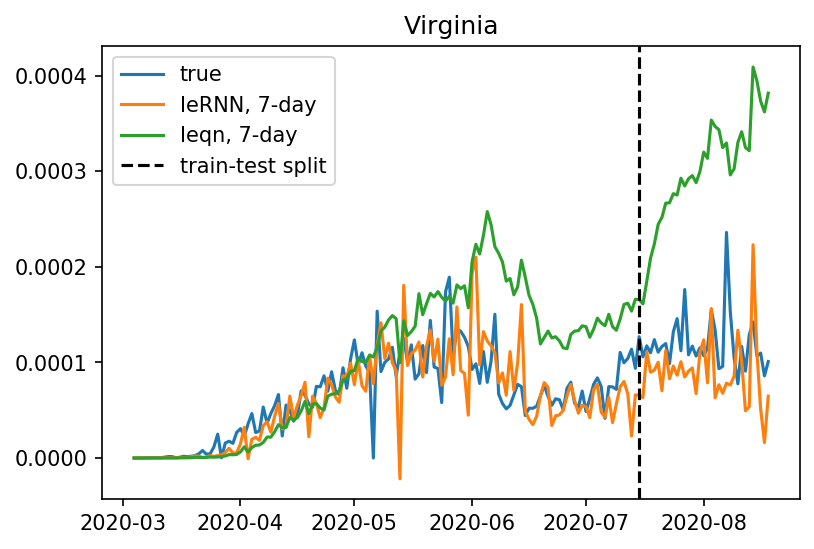}
}
\caption{Comparing 7-day ahead predictions of IeRNN and I-equation with training (testing) period to the left (right) of the vertical dashed line.  The vertical axis is fraction of newly infected people in the population. The horizontal axis is time in unit of days.}
\label{fig:7day}
\end{figure}

Model robustness in training is an important attribute, so that the model performance is not sensitive to initialization of parameters ($\alpha$,$\beta_1$,$\beta_2$,$\gamma$,$\sigma_1$,$\sigma_2$) during training. 
We find that the I-equation (I-model with $I_e=0$)  is not easy to learn in the sense that a sub-optimal local minimum is often reached by gradient descent during optimization. With coupling to RNN ($I_e \not = 0$) in IeRNN, the landscape of loss function is regularized so that a local minimum from any random initialization gives a robust and accurate fit. 
Fig. \ref{fig:1day} shows that I-equation is much less accurate in 1-day ahead prediction than IeRNN. Fig. \ref{fig:7day} illustrates the same outcome in 7-day ahead prediction.

In further experiment, we train and test IeRNN and I-equation with randomly initialized parameters ($\alpha$,$\beta_1$,$\beta_2$,$\gamma$,$\sigma_1$,$\sigma_2$). By repeating the training and testing procedure for 20 times, we compare the average MSE loss for both models. The results in Tables \ref{tab:1day} and  \ref{tab:7day} show that IeRNN performs better for both training loss and testing loss in 1-day ahead and 7-day ahead predictions. 

\begin{table}[htbp]
    \centering
    \begin{tabular}{c c|c c}
        \hline \hline
        & & IeRNN & I-equation\\
        \hline
        \multirow{2}{*}{California} & training & 7.63e-09 & 8.49e-08 \\
        & testing & 1.26e-08 & 9.68e-07 \\
        \multirow{2}{*}{Florida} & training & 4.24e-08 & 3.45e-06 \\
        & testing & 3.59e-08 & 3.97e-05 \\
        \multirow{2}{*}{Virginia} & training & 3.70e-09 & 2.60e-08 \\
        & testing & 6.90e-09 & 1.56e-07 \\
        \hline \hline
    \end{tabular}
    \caption{Average MSE's of training (testing) loss in 1-day ahead prediction.}
    \label{tab:1day}
\end{table}

\begin{table}[htbp]
    \centering
    \begin{tabular}{c c|c c }
        \hline \hline
        & & IeRNN & I-equation\\
        \hline
        \multirow{2}{*}{California} & training & 8.01e-09 & 1.32e-07 \\
        & testing & 9.66-09 & 1.62e-06 \\
        \multirow{2}{*}{Florida} & training & 8.15e-09 & 1.49e-06 \\
        & testing & 9.77e-09 & 2.20e-05 \\
        \multirow{2}{*}{Virginia} & training & 7.69e-09 & 8.21e-08 \\
        & testing & 2.03e-08 & 1.41e-06 \\
        \hline \hline
    \end{tabular}
    \caption{Average MSE's of training (testing) loss in 7-day ahead prediction.}
    \label{tab:7day}
\end{table}

\subsection{1-day ahead prediction}
We compare IeRNN (with $\beta_1(t)$), IeRNN, LSTM and ARIMA on 1-day ahead prediction. 
IeRNN achieves lower MSE error than LSTM and ARIMA on test set. With policy response function $\beta_1(t)$, IeRNN gives further improvement beyond IeRNN with constant $\beta_1$, see Table 3.

\begin{table}[htbp]
    \centering
    \begin{tabular}{c | c c c c}
        \hline \hline
        & IeRNN & IeRNN & LSTM & ARIMA\\
        & $\beta_1$(t) & & & \\
        \hline
        California & 1.83e-09 & 2.45e-09 & 5.00e-09 & 1.44e-08\\
        Florida & 6.13e-09 & 7.55e-09 & 4.68e-08 & 4.11e-08\\
        Virginia & 1.27e-09 & 1.29e-09 & 3.37e-09 & 3.74e-09\\
        \hline \hline
    \end{tabular}
    \caption{MSE comparison of different models on 1-day ahead prediction.}
    \label{tab:testcompare1day}
\end{table}

\begin{figure}
\subfigure{
    \includegraphics[width=0.5\textwidth]{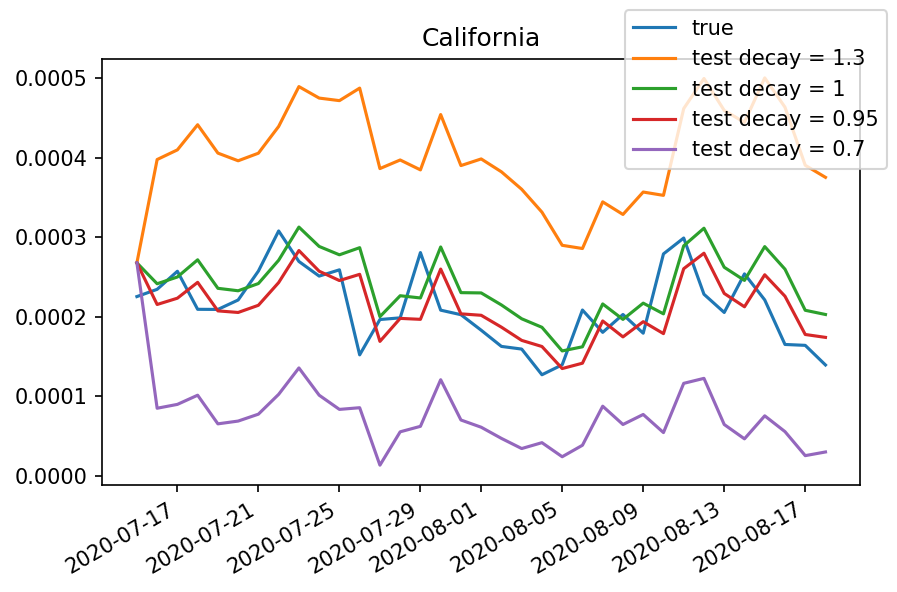}
}
\subfigure{
    \includegraphics[width=0.5\textwidth]{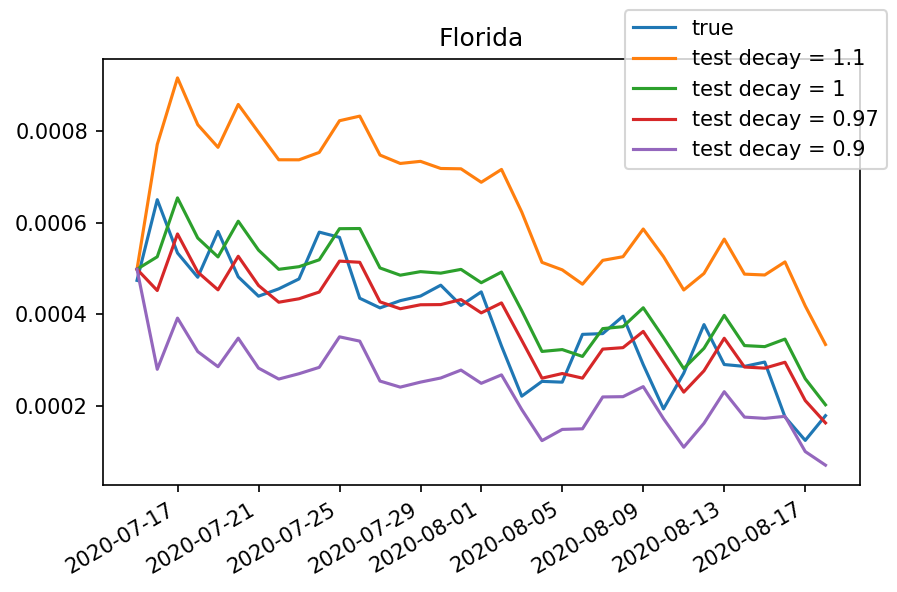}
}
\subfigure{
    \includegraphics[width=0.5\textwidth]{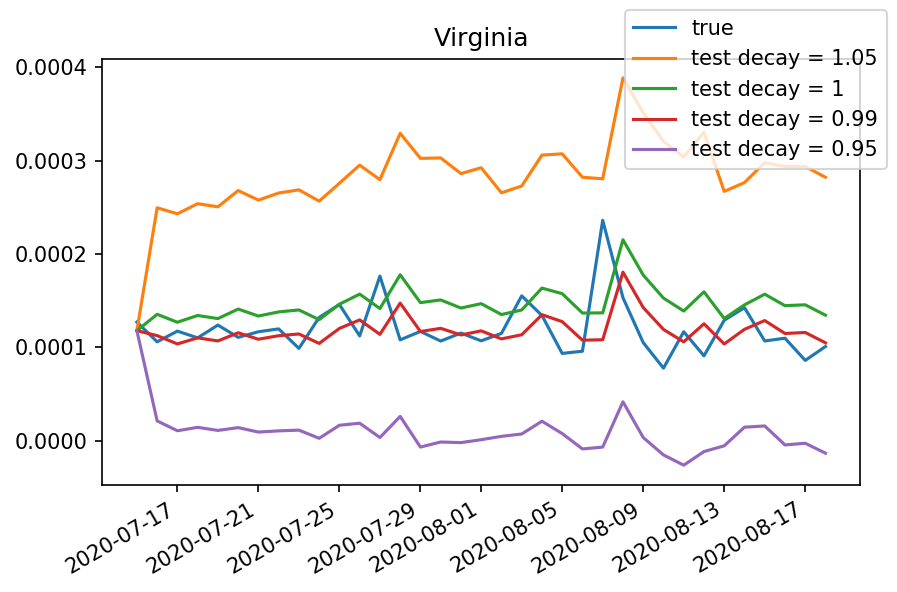}
}
\caption{Effect of test decay (policy response multiplier) in test period of 1-day ahead prediction task. The IeRNN is trained through March 3, 2020 to July 14, 2020. The vertical axis is fraction of newly infected people in the population. The horizontal axis is time in unit of days. }
\label{fig:testdecay1}
\end{figure}

\begin{figure}
\subfigure{
    \includegraphics[width=0.5\textwidth]{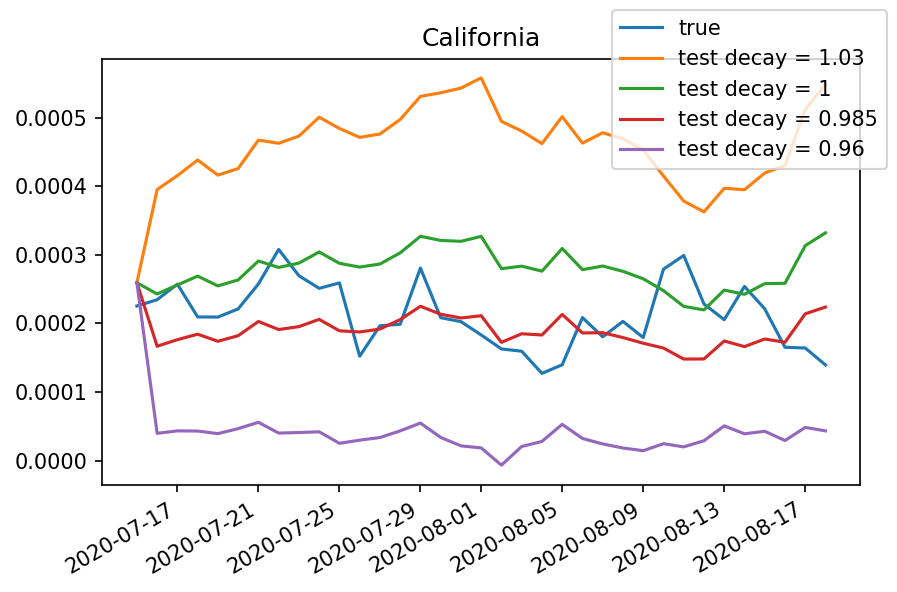}
}
\subfigure{
    \includegraphics[width=0.5\textwidth]{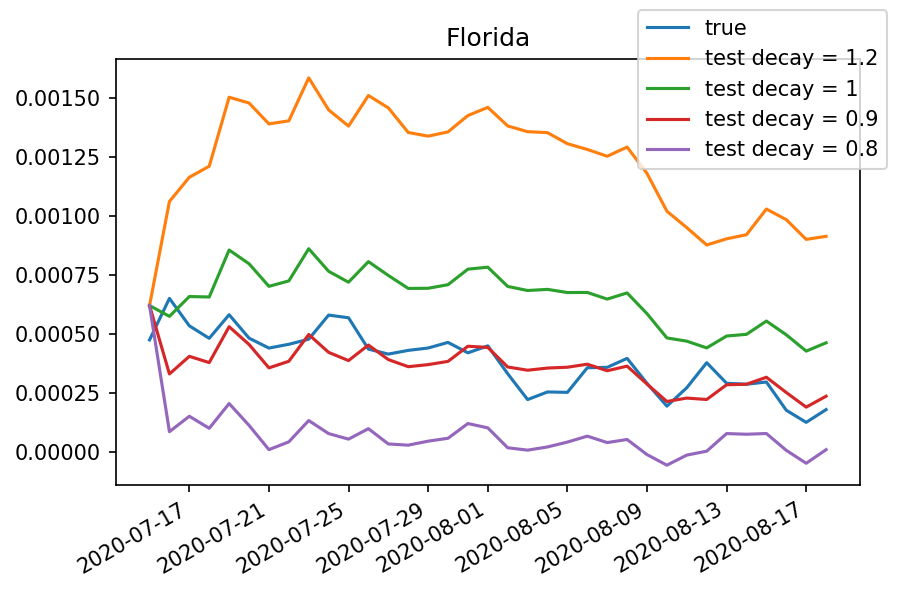}
}
\subfigure{
    \includegraphics[width=0.5\textwidth]{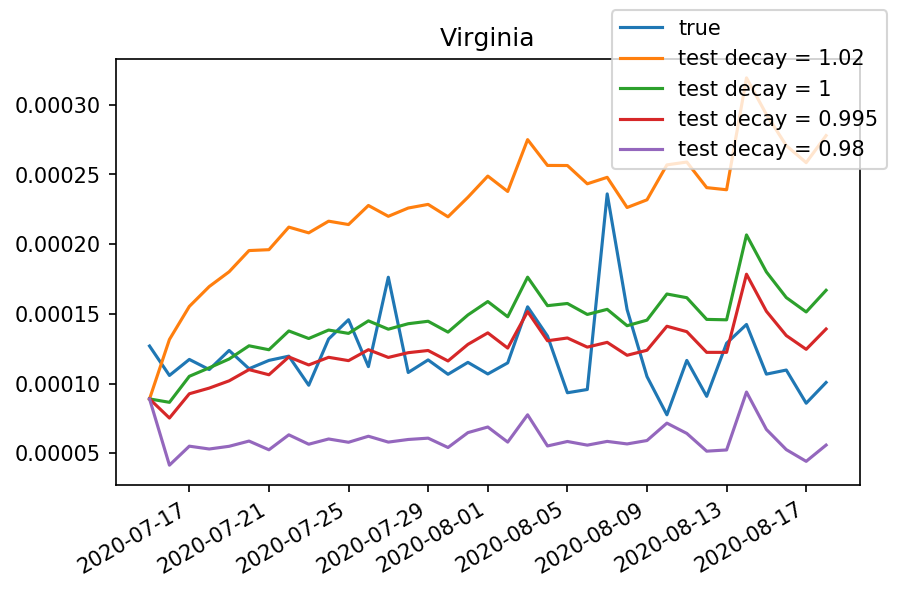}
}
\caption{Effect of test decay (policy response multiplier) in test period  of 7-day ahead prediction task. The IeRNN is trained through March 3, 2020 to July 14, 2020. The vertical axis is fraction of newly infected people in the population. The horizontal axis is time in unit of days. }
\label{fig:testdecay7}
\end{figure}

\subsection{7-day ahead prediction}
Motivated by weekly forecasting from CDC, we study the 7-day ahead prediction task. 
The loss function is modified by replacing the I-model by a 7-day delayed version below:

\begin{dmath}
  \hat{I}_{t} =  \sigma_1 \alpha + (1 - \sigma_1 - \gamma) I_{t-7} + \sigma_2 I_{e, t-7} - \gamma \frac{t - t_0}{P + 1}  \sum_{j = 0}^P I_{t-7-j} - S_0 \exp \left( - \frac{t - t_0}{P + 1}  \sum_{j = 0}^P (\beta_1 I)_{t-7-j} + (\beta_2 I_{e})_{t-7-j} \right)
\label{eq:7dayloss}
\end{dmath}
where the output value at time $t$ is influenced by the feature vector $I_e$ at time $t-7$ and later. 
With a similar 
modification of loss function,  
we  adapt LSTM to the 7-day ahead prediction. Table 4 compares IeRNN and LSTM in terms of MSE on testing data.

\subsection{Effect of Policy Response $\beta_1 (t)$ in Testing}
To study the effect of policy response in IeRNN model on testing data, we multiply the learned $\beta_1 (t)$ by a constant factor (called \textit{test decay}) during testing. 
Fig. \ref{fig:testdecay1} and Fig. \ref{fig:testdecay7} show the impact 
to model prediction on test data by adjusting \textit{test decay} which  could control the future trend of infection.

\begin{table}[htbp]
    \centering
    \begin{tabular}{c | c c c c}
        \hline \hline
        & IeRNN & IeRNN & LSTM \\
        & $\beta_1(t)$ & & \\
        \hline
        California & 6.79e-09 & 9.84e-09 & 1.49e-08 \\
        Florida & 4.34e-08 & 4.47e-08 & 5.74e-08 \\
        Virginia & 1.16e-09 & 1.55e-09 & 1.54e-08 \\
        \hline \hline
    \end{tabular}
    \caption{MSE comparison of different models on 7-day ahead prediction.}
    \label{tab:testcompare7day}
\end{table}

\section{Conclusions}
We develop a novel spatio-temporal infectious disease model called IeRNN, which is a hybrid model consisting of I-equation from SEIR driven by spatial features. With such features and RNN dynamics as external input to the I-equation, the robustness to parameter initialization in model training is greatly improved. In 1-day and 7-day ahead prediction, our model outperforms standard temporal models. In future work, the social control mechanisms \cite{Pareschi_2020,Levin_2020} could be considered to strengthen the I-equation, as well as traffic data to expand inflow effect beyond geographic neighbors.

\section{Acknowledgement}
The work was partially supported by NSF grants IIS-1632935, DMS-1924548.

\bibliography{main.bib}

\end{document}